\documentclass[sigconf,nonacm]{aamas}

\usepackage{balance}
\usepackage{tikz}
\usetikzlibrary{positioning}
\graphicspath{{figures/}}
\newcommand{\aitoolversion}{Claude Opus 5.5}
\newcommand{\jlens}{J-lens}
\newcommand{\eo}{e_2}
\newcommand{\ep}{e_2'}

\acmConference[Preprint]{Preprint}{October 2026}{}

\title{Silent Dissent: LLM Agents That Yield to the Majority Still Represent Their Original Premise}

\author{Ziang Ni}
\authornote{Both authors contributed equally to this research.}
\affiliation{
  \institution{Delft University of Technology}
  \city{Delft}
  \country{The Netherlands}}
\email{ziangni@tudelft.nl}
\author{Peng Zou}
\authornotemark[1]
\affiliation{
  \institution{Sun Yat-sen University}
  \city{Guangzhou}
  \country{China}}
\email{zoup7@mail2.sysu.edu.cn}

\begin{abstract}
Multi-agent debate is increasingly used to reach consensus among LLM agents, yet agents often yield to a unanimous majority. When an agent changes its answer, has it changed its mind or only its statement? We study this with two-hop factual questions whose intermediate entity (the bridge, e.g.\ the country in ``the capital of the country where the Sagrada Familia is located'') is never stated by anyone. Scripted peers, in the role of Asch's confederates, unanimously assert a wrong answer taken from another fact with a different bridge. At the moment the agent answers, we read the bridge from its residual stream with the Jacobian lens (J-lens) and, for comparison, the logit lens. In pre-registered tests on held-out facts with four open-weight models, agents of Qwen3.5-4B, Qwen3.6-27B and Gemma-4-E4B-it that gave in still represented their original bridge in the pre-registered layers below the output (hit@100 above a control entity: 0.85, 0.22 and 0.24), where the logit lens rarely ranked it among the top 100 tokens (0.00--0.06). These agents also represented the bridge behind the peers' answer, beyond a mention baseline. A pre-registered addendum hid the agent's earlier answer or removed it altogether: agents that gave in still represented their original bridge in all four models (0.43, 0.29, 0.37 and 0.25 with the answer hidden), including Llama-3.1-8B-Instruct, which barely did so with its answer in view (0.03). The premise therefore does not require the agent's own earlier answer; it can be computed from the question alone while the agent states the majority's answer. Hiding the earlier answer also changed conformity: Qwen3.5-4B gave in on 89\% of questions instead of 8\%. In exploratory interventions, injecting the bridge's J-lens direction brought agents back to their original answer only in the two Qwen models. Stated consensus in multi-agent debate can thus overstate agreement. We also report the negative results of our pre-registered program.
\end{abstract}


\begin{document}

\pagestyle{fancy}
\fancyhead{}

\maketitle
\thispagestyle{empty}

\section{Introduction}

In multi-agent debate, several LLM agents exchange answers over rounds until they converge; the protocol is used to improve factuality and reasoning~\cite{du2024debate,liang2024encouraging}. Convergence is not the same as persuasion. LLM agents yield to unanimous peers much as people do in Asch's experiments~\cite{asch1956studies,zhu2025conformity,weng2025benchform,zhang2024exploring}; debate tends to settle on the majority view even when that view is a shared mistake~\cite{estornell2024multillm}; and much of what debate adds over a single agent can be traced to majority voting~\cite{choi2025debate}. When an agent abandons a correct answer to join the majority, has it changed its mind, or only its statement?

Social psychology separates public compliance from private acceptance~\cite{kelman1958compliance,deutsch1955study}, and the gap between them, preference falsification~\cite{kuran1995private}, matters for any group that aggregates what its members say. If agents only say what the majority says, the stated consensus of a debate overstates agreement, and a count of converted agents mixes persuaded agents with agents that went along. For LLM agents, a proxy for the private side can be read from activations.

Where to read it is not obvious. We first tried the most direct place: the answer itself, a multiple-choice letter, read where the agent states it. A positive control failed. When a coordinator told the agent which letter to report regardless of its own answer, a readout at a layer fixed in advance ranked the agent's own letter first in 0 of 1{,}000 cases, although the stated and the preferred answer were known to differ. Recent mechanistic studies of sycophancy that read answer tokens likewise find the correct answer overridden or erased in late layers~\cite{wang2026truth,joswin2026authority}.

\begin{figure}[t]
\centering
\begin{tikzpicture}[font=\scriptsize,
  box/.style={draw=black!40, rounded corners=2pt, align=left, text width=0.93\columnwidth, inner sep=3pt}]
\node[box, fill=black!4] (r0) {\textbf{Round 0.} Question: Complete the statement: The capital city of the country where the Sagrada Familia is located is \dots\\
\textbf{Agent 1:} The capital city of the country where the Sagrada Familia is located is Madrid.};
\node[box, fill=orange!8, below=3pt of r0] (r1) {\textbf{Round 1.} \textbf{Agents 2--4} (each): The capital city of the country where the Sagrada Familia is located is Rome.\\
\textbf{Moderator:} Taking the other agents' statements into account, Agent 1, complete the statement again.\\
\textbf{Agent 1:} The capital city of the country where the Sagrada Familia is located is \fbox{\strut\,?\,}};
\node[box, fill=blue!5, below=3pt of r1] (rd) {\textbf{Stated:} Rome \emph{(gave in)}.\\
\textbf{Read at \fbox{\,?\,}} (last prompt token, pre-registered layers), each relative to a control bridge that appears nowhere (e.g.\ \emph{Japan}): \emph{Spain} (original bridge $\eo$) and \emph{Italy} (peers' bridge $\ep$) rank high under the \jlens{}; under the logit lens they mostly do not.};
\end{tikzpicture}
\caption{The paradigm (illustrative, not a recorded transcript; plain format, abridged). Neither bridge is written anywhere in the transcript. Peers can instead attack the first hop (``The country where the Sagrada Familia is located is Italy. The capital city of Italy is Rome.'') or the second hop (right bridge, wrong capital).}
\label{fig:paradigm}
\Description{Transcript of a two-round debate and the lens readout at the agent's answer position.}
\end{figure}
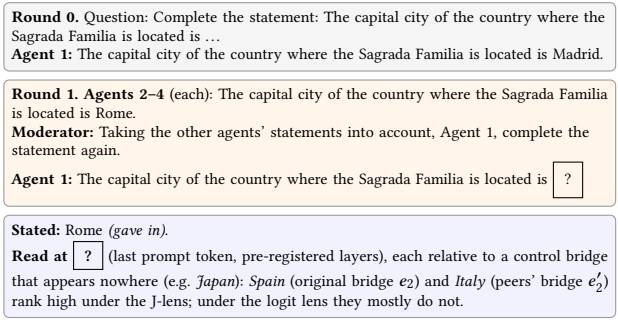

We therefore read a premise instead of the answer. We use two-hop factual statements whose intermediate entity, the \emph{bridge}, is never stated by anyone (Figure~\ref{fig:paradigm}). The agent first completes the statement alone. Scripted peers, in the role of Asch's confederates, then unanimously assert a wrong answer, taken from another fact with a different bridge. At the position where the agent begins its second answer, we decode the residual stream with the Jacobian lens (\jlens{})~\cite{gurnee2026verbalizable}, which reads what a model is poised to verbalize, and, for comparison, with the logit lens~\cite{nostalgebraist2020logitlens}. We call an agent that gave in but still represents its original bridge a \emph{silent dissenter}: ``silent'' refers to what it states, and ``dissent'' to a represented premise, not a belief. In agents that gave in, the \jlens{} read the original bridge 0.85, 0.22 and 0.24 above a control entity in three of four models, where the logit lens rarely ranked it among the top 100 tokens (at most 0.06). A pre-registered addendum shows that this does not require the agent to read the bridge back from its own earlier answer: with that answer hidden or removed, agents that gave in still represent the bridge in all four models.

We ran the study as a pre-registered program~\cite{nosek2018preregistration}: for each model, the readout layers, categories and hypotheses were fixed on a development split and committed before any of its test data were collected. Our contributions are:
\begin{itemize}
\item A paradigm that separates an agent's stated answer from a premise it never states, tested with pre-registered hypotheses on held-out facts for Qwen3.5-4B, Qwen3.6-27B, Gemma-4-E4B-it and Llama-3.1-8B-Instruct.
\item Evidence of silent dissent in three of the four models: agents that gave in still represent their original bridge and the bridge behind their peers' answer. In intermediate layers the \jlens{} brings the original bridge into the top 100 tokens where the logit lens does not, also for Gemma-4-E4B-it, whose layers were chosen without the logit lens; rank differences below that threshold depend on the model.
\item A pre-registered control that hides or removes the agent's earlier answer: agents that gave in still represent the bridge in all four models, including Llama, which retains almost nothing with the answer in view; hiding it also changes how often agents give in (Qwen3.5-4B: 8\% to 89\%).
\item Exploratory analyses: injecting the bridge's direction restores the original answer in the two Qwen models but not in Gemma, whose bridge is about as readable; in free debate, Llama agents that gave in still represent their bridge; and the program's negative results.
\end{itemize}

\section{Related Work}

\paragraph{Debate and aggregation.}
Multi-agent debate improves factual accuracy and reasoning in some settings~\cite{du2024debate,liang2024encouraging}, but its gains depend on configuration~\cite{smit2024mad}, much of them come from majority voting~\cite{choi2025debate}, and debate among similar agents converges to the majority even when the majority is wrong~\cite{estornell2024multillm}. Voting helps when competent voters err independently~\cite{list2001epistemic}; social influence erodes that independence and the diversity of answers without improving the collective estimate~\cite{lorenz2011social}. Opinion-dynamics models in which each agent holds an expressed and a private opinion reproduce Asch's results~\cite{ye2019influence}. We read a representational proxy for the private side of LLM agents from their activations.

\paragraph{Conformity and sycophancy in LLMs.}
Asch-style experiments show that LLMs conform to unanimous wrong majorities~\cite{zhu2025conformity,weng2025benchform}, more so when uncertain~\cite{zhu2025conformity}; agent societies show conformity-like dynamics~\cite{zhang2024exploring}; models agree with users against their own knowledge~\cite{sharma2024sycophancy}, and multi-turn persuasion flips their factual answers~\cite{xu2024earth}. A behavioural study assigns agents private opinions that oppose a group norm and finds that most of them conform in public~\cite{yashwanth2026pluralistic}. They measure what agents say; we read what they represent.

\paragraph{Model internals under social pressure.}
Several recent studies inspect model internals when an LLM yields. With a single user or an authority figure, logit-lens and probing analyses of multiple-choice answers find the correct answer overridden or erased in late layers~\cite{wang2026truth,joswin2026authority}, or attention heads that still signal that the endorsed claim is wrong~\cite{pandey2026sycophancy}. Under simulated peer disagreement, activation patching localizes the damage to a mid-layer window~\cite{kumarappan2026notjust}. Models adopt external candidate answers (from tools, retrieval, other agents or users) even after rejecting them when asked to check them, and a \jlens{} decomposition separates that verification from the adoption~\cite{kawada2026evidence}; with the \jlens{}, a single model that goes along with a false presupposition still represents the true entity~\cite{davide2026endognostics}. Most of these studies read the answer or a truth signal under one source of pressure. We read a premise that no one ever states, under a unanimous majority; we compare the \jlens{} with the logit lens in pre-registered held-out tests on four models; and we test whether the premise depends on the agent seeing its own earlier answer.

\paragraph{Represented but unsaid.}
Models can represent what they do not output: correct labels survive in intermediate layers under false demonstrations until late layers override them~\cite{halawi2024overthinking}, unsupervised probes recover truth even when models are prompted to answer falsely~\cite{burns2023discovering}, and probed world states stay faithful while outputs do not~\cite{feng2025monitoring}. The logit lens~\cite{nostalgebraist2020logitlens} and the tuned lens~\cite{belrose2023tuned} decode intermediate layers into the vocabulary; the \jlens{} refines the logit lens with the average Jacobian from each layer to the output, reads intermediate steps that are never verbalized, including the bridges of two-hop prompts, and steers them through lens directions; the tuned lens, by contrast, tends to skip ahead to the output~\cite{gurnee2026verbalizable}. Two-hop facts are the standard testbed for such latent steps: models often resolve the bridge internally~\cite{yang2024latent}, sometimes too late for the second hop~\cite{biran2024hopping}, and latent composition succeeds far more often for country than for year bridges~\cite{yang2025shortcuts}. We follow~\cite{gurnee2026verbalizable} in reading and injecting bridges; what we add is the social-pressure paradigm, the pre-registered cross-model test and the own-answer control.

\section{Method}

\subsection{Facts, agents and conditions}
\paragraph{Facts.} TwoHopFact~\cite{yang2024latent} contains 45{,}595 statements that compose a first hop $e_1 \to \eo$ with a second hop $\eo \to e_3$ (e.g.\ ``The author of the novel $e_1$ was born in the city of \dots''); we add a small set of hand-written capital-city items. A model \emph{knows} a fact if it completes the first hop, the second hop and the composed statement correctly; this holds for 4.5\% of facts for Qwen3.5-4B, 8.7\% for Qwen3.6-27B, 3.2\% for Gemma-4-E4B-it and 9.2\% for Llama-3.1-8B-Instruct. Known facts are split into development and test sets by bridge entity (30\% of bridges to development), so no bridge appears in both. Each fact is paired with a \emph{peer} fact (bridge $\ep$, answer $e_3'$) and a \emph{control} fact: other known facts of the same category, drawn at random per item, whose bridges and answers differ from the item's and from each other. The peers thus argue for the answer of another true fact, and the control bridge appears in no transcript. For capital items the peers' answer determines their bridge (Rome implies Italy); in the other categories it fits many bridges (in TwoHopFact, 96--98\% of the answers in the birth-city, university-city and university-country categories belong to more than one bridge), so there the peers' particular bridge is associated with their answer rather than implied by it. The bridge never appears in the transcript, except that some university names contain their city (7\% of the university-city category in TwoHopFact).

\paragraph{Models.} Qwen3.5-4B~\cite{qwen2026qwen35} was the primary model; Qwen3.6-27B~\cite{qwen2026qwen36} and Llama-3.1-8B-Instruct~\cite{grattafiori2024llama3} were planned replications across scale and family, and Gemma-4-E4B-it~\cite{gemmateam2026gemma4} adds a further family. Every registered test that was run is reported here.

\paragraph{Agents.} The agent under study (Agent~1) and $n \in \{1,3,5\}$ scripted peers take part in a plain-text transcript that opens with a two-sentence header (``The following is a transcript of AI agents completing factual statements together. Each agent gives its own answer.'') and two worked examples; the model continues the transcript in Agent~1's voice. In round~0, Agent~1 completes the composed statement alone. In round~1, its own statement stays in context, the peers speak, a moderator asks Agent~1 to complete the statement again, and the prompt ends with the statement prefix, so the next token begins its answer (Figure~\ref{fig:paradigm}). Answers are greedy 12-token completions; an answer counts as original, as the peers' answer (the agent \emph{gave in}) or as other if its normalized text begins with an alias of $e_3$, of $e_3'$, or neither. Only items answered correctly in round~0 enter round~1 (668, 476, 476 and 542 test items). For Llama the selection rule chose its chat template on the development split: a system message replaces the header, there are no worked examples, the agent's statements are assistant turns, and each round's peer statements and moderator request form one user turn.

\paragraph{Conditions.} Under \emph{answer pressure} all peers state the composed statement with $e_3'$; under \emph{first-hop pressure} they state the wrong bridge $\ep$ and its second hop to $e_3'$; under \emph{second-hop pressure} they state the correct bridge $\eo$ and a wrong second hop to $e_3'$. Baselines: in \emph{agree} all peers state $e_3$; in \emph{mention} they agree but the first says it ``briefly considered'' $e_3'$; in \emph{mention-bridge} they agree but the first reports that ``some say'' the bridge is $\ep$. On the development split two further conditions validate the readout: \emph{hypothetical} (the moderator asks the agent to assume the bridge is $\ep$) and \emph{instructed} (a coordinator requires every agent to state $e_3'$ ``regardless of what they believe''). An addendum repeats round~1 with the agent's round-0 statement replaced by ``(answer submitted)'' (\emph{hidden}) or left out, so that the agent answers for the first time after the peers and the moderator asks it to ``complete the statement'' (\emph{absent}).

\subsection{Reading the bridge}
Let $h_\ell$ be the residual stream at layer $\ell$ (numbered from 0) at the last prompt token, where the agent's answer begins. The logit lens decodes $W_U\,\mathrm{norm}(h_\ell)$; the \jlens{} decodes $W_U\,\mathrm{norm}(J_\ell h_\ell)$ with $J_\ell = \mathbb{E}[\partial h_L / \partial h_\ell]$, the input--output Jacobian averaged over positions of a generic corpus~\cite{gurnee2026verbalizable}. For the Qwen and Llama models we use the pre-fitted lenses hosted by Neuronpedia~\cite{neuronpedia} (files and revision in the supplement). Its Gemma lens was fitted on the base model, so our primary Gemma lens is refitted on the instruction-tuned model from 100 WikiText sequences with a port of the reference estimator (it matches the reference code to $10^{-8}$ on a shared model); the base-model lens is a secondary readout.

An entity is probed by the first token of each of its content words (at least three letters, excluding stop words such as \emph{University} or \emph{Republic}); its rank is the best full-vocabulary rank among them. Tokens shared by two roles of the same item are removed, and a record left with no token for a role is dropped from statistics on that role. A \emph{record} is one turn of Agent~1: one item in one condition and round. For a role $r$ (original bridge $\eo$ or peer bridge $\ep$) the readout of a record is $\mathrm{hit@}k(r) - \mathrm{hit@}k(\text{control bridge})$, averaged over the layers of the pre-registered band, with $k = 100$, the top 0.04--0.08\% of vocabularies of 128{,}256 to 262{,}144 tokens; a statistic is its mean over records. The last layer is excluded from every band: it is the model's output and identical under both lenses.

\subsection{Pre-registration}\label{sec:prereg}
For each model, the prompt format, the layer band, the categories and the validation checks were chosen on the development split only and written to a pre-registration file that was committed before any of that model's test runs and never edited. Rules changed between models, always on the new model's development data, and every change is recorded in the files. On Qwen3.5-4B's development split the declared threshold $k = 10$ selected no layers, so $k = 100$ was adopted before any test run (the control's own hit@100 in the band is 0.02--0.05) and kept for all models. Two band rules were used.
\emph{Rule v1}: the band is every layer (except the last) where correctly answered round-0 records read the original bridge at $\geq 0.5$; categories need a band mean $\geq 0.5$; an adoption check requires the peer bridge to read at least 0.3 higher when the agent is asked to assume it (hypothetical) than when it is only mentioned (mention-bridge), and the original bridge to persist when peers agree (at least half its round-0 value).
\emph{Rule v2}: categories are those where the \jlens{} exceeds the logit lens by $\geq 0.25$ in round~0; the band is every layer where, in agree dialogues, the \jlens{} reads $\geq 0.3$ and exceeds the logit lens by $\geq 0.25$; the adoption check becomes a ratio ($\geq 3\times$), with the same persistence check.
Rule v1 was set on Qwen3.5-4B. On Qwen3.6-27B it selected only layer 62 of 64, where both lenses read the bridge equally on the development split, which would have left the lens comparison untestable; rule v2 was written on 27B's development data before its test run, and the v1 file was kept. The change of the adoption check from a gap to a ratio relaxed it: under the absolute gap, Qwen3.6-27B's development data would have failed (0.28 < 0.30), which the registration records. Llama-3.1-8B-Instruct failed v1 (its bridge readout stayed below the threshold, and the \jlens{} did not beat the logit lens) and passed v2 in chat format. Rule v2 builds a \jlens{} advantage into the band, but it is measured on development items, whose bridges never occur in the test set, and on agents that answered alone or whose peers agreed; H4 is tested on test agents that gave in. For the v2 models H4 therefore asks whether the advantage survives giving in, not whether it exists. Gemma-4-E4B-it selected nothing under v2, because the bridge appears in both lenses at once near the output; it was registered under v1 unchanged, and its registration predicted that H4 would fail. Section~\ref{sec:robust} reports each model under the other rule where that rule selects a band.

The hypotheses concern round~1 with three peers under answer pressure (H3: first- and second-hop pressure):
\textbf{H1} (retention): agents that gave in read their original bridge above control.
\textbf{H2} (registered as ``implied premise''): agents that gave in read the peers' bridge higher than agents that kept their answer in the mention condition.
\textbf{H3} (locus): among agents that gave in, the original bridge reads higher under second-hop than under first-hop pressure (in hindsight a check on the conditions; \S\ref{sec:h3}).
\textbf{H4} (lens): the H1 statistic is larger under the \jlens{} than under the logit lens (paired within records).
A hypothesis is supported if the lower end of its 95\% percentile bootstrap interval~\cite{efron1993introduction} is above zero (1{,}000 resamples of records; 2{,}000 for the two-sample H2 and H3). Items that share a bridge are not independent; Section~\ref{sec:robust} adds intervals clustered by bridge and a correction for multiple tests.
The addendum was registered after development pilots that checked only the answer format and the agree readout, and before its test runs; readout, format and categories are those of each model's registration. For agents that gave in (answer pressure, three peers) it predicts: \textbf{O1}, the original bridge reads above control under hidden; \textbf{O3}, also under absent; \textbf{O2}, higher under hidden than under absent (a contribution of having answered); \textbf{O4}, higher under the \jlens{} than under the logit lens under hidden. All other analyses are exploratory.

\begin{table}[t]
\caption{Setup on the test split. Band\,/\,$L$: pre-registered layers / number of layers. Items: test items / items entering round~1 (answered correctly in round~0). Gave in: share of agents stating the peers' answer in round~1 with three peers. Categories (first entity $\to$ bridge $\to$ answer): L landmark $\to$ country $\to$ capital; B person $\to$ birth city $\to$ country; U university $\to$ city $\to$ country; P$_c$, P$_n$ person $\to$ undergraduate university $\to$ its city or country.}
\label{tab:setup}
\small
\setlength{\tabcolsep}{2.1pt}
\begin{tabular}{lllllrrr}
\toprule
 & & & & & \multicolumn{3}{c}{Gave in (\%)} \\
\cmidrule(lr){6-8}
Model & Rule & Band\,/\,$L$ & Cat. & Items & ans. & hop 1 & hop 2 \\
\midrule
Qwen3.5-4B & v1 & 23--30\,/\,32 & L, B, U & 718/668 & 8 & 8 & 0 \\
Qwen3.6-27B & v2 & 52--55\,/\,64 & L, B, P$_c$ & 514/476 & 39 & 49 & 6 \\
Gemma-4-E4B & v1 & 32, 34--40\,/\,42 & L, B, U & 533/476 & 25 & 52 & 24 \\
Llama-3.1-8B$^\ast$ & v2 & 17--23\,/\,32 & L, B, P$_n$ & 570/542 & 90 & 97 & 8 \\
\bottomrule
\end{tabular}
\\[2pt]
\raggedright\footnotesize $^\ast$Chat format; all other models use the plain transcript. Gemma and Llama are the instruction-tuned models (-it, -Instruct).
\end{table}

\section{Results}

\subsection{How often agents give in}
Three unanimous peers asserting a wrong answer made 8\% of Qwen3.5-4B's agents give in, 25\% of Gemma's, 39\% of Qwen3.6-27B's and 90\% of Llama's (Table~\ref{tab:setup}). Attacking the first hop, i.e.\ stating a wrong bridge, persuaded about as often (Qwen3.5-4B) or more; attacking the second hop while stating the correct bridge persuaded least (for Gemma only slightly: 24\% vs.\ 25\%). This ordering holds for three peers; with one or five peers it differs for some models. A single peer rarely moved an agent under answer pressure (0--10\%), and more peers did not always persuade more: Gemma gave in to three peers in 25\% of cases but to five in 8\%. These comparisons are exploratory, without intervals, and the kinds of pressure also differ in length and content.

\begin{table*}[t]
\caption{Pre-registered hypotheses on the test split: estimate, 95\% bootstrap interval and number of records (H2, H3: agents that gave in $\times$ comparison group). H1: original-bridge readout of agents that gave in; H2: their peer-bridge readout minus that of mention-condition agents; H3: original-bridge readout under second- minus first-hop pressure; H4: H1 under the \jlens{} minus under the logit lens, paired. Every interval excludes zero. $^\dagger$Band chosen by rule v2 for a \jlens{} advantage on development data, so H4 tests whether the advantage persists in test agents that gave in. For Qwen3.6-27B the logit lens reads 0.000 on these records, so H4 equals H1. With intervals clustered by bridge and a Holm correction, all but Llama's H2 remain supported (\S\ref{sec:robust}).}
\label{tab:hyp}
\small
\begin{tabular}{lllll}
\toprule
Model & H1 retention & H2 peers' bridge & H3 hop 2 $-$ hop 1 & H4 \jlens{} $-$ logit lens \\
\midrule
Qwen3.5-4B & 0.852 [0.782, 0.917], $n{=}54$ & 0.373 [0.261, 0.491], $55{\times}661$ & untestable ($0{\times}53$) & 0.792 [0.722, 0.857] \\
Qwen3.6-27B$^\dagger$ & 0.223 [0.174, 0.280], $n{=}168$ & 0.126 [0.080, 0.173], $172{\times}448$ & 0.323 [0.115, 0.533], $12{\times}213$ & 0.223 [0.174, 0.280] \\
Gemma-4-E4B-it & 0.237 [0.175, 0.299], $n{=}120$ & 0.211 [0.163, 0.258], $120{\times}475$ & 0.555 [0.476, 0.631], $112{\times}244$ & 0.192 [0.137, 0.248] \\
Llama-3.1-8B-Instruct$^\dagger$ & 0.034 [0.020, 0.048], $n{=}485$ & 0.027 [0.014, 0.041], $487{\times}540$ & 0.400 [0.260, 0.533], $44{\times}521$ & 0.030 [0.018, 0.043] \\
\bottomrule
\end{tabular}
\end{table*}

\begin{figure*}[t]
\centering
\includegraphics[width=\textwidth]{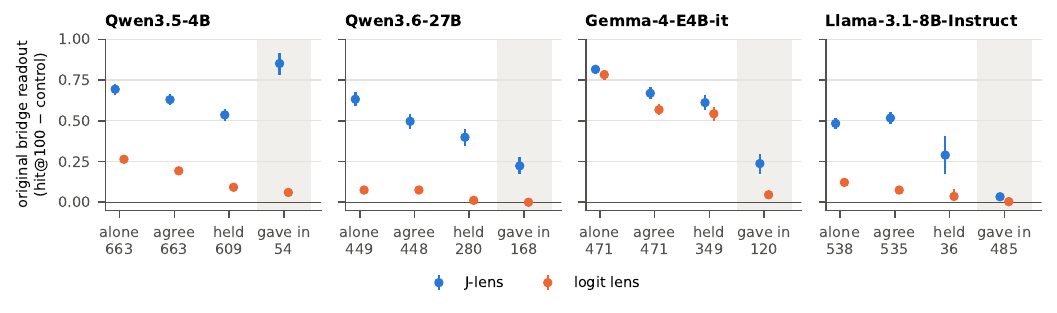}
\caption{Original-bridge readout in the pre-registered band (hit@100 minus control; mean, 95\% CI, records under each label) for agents answering alone in round~0, agents whose peers agreed, agents that held under answer pressure, and agents that gave in. Round~0 and agree include all items; held and gave in split the pressure items by outcome, so the conditions are different item sets. Within each condition the two lenses are compared on the same records.}
\label{fig:conditions}
\Description{Four panels, one per model, with J-lens and logit-lens readouts across four conditions.}
\end{figure*}

\subsection{Silent dissent (H1, H4)}
H1 and H4 compare within records: the original bridge with a control entity, and the \jlens{} with the logit lens, for the same agents at the same position. Comparisons across conditions (Figure~\ref{fig:conditions}) are between item sets: held and gave in split the pressure items by outcome, and models give in more on items they are unsure of~\cite{zhu2025conformity}, so such differences may partly reflect which items flip.

Agents that gave in still represented their original bridge, and in the pre-registered layers the \jlens{} read it where the logit lens read little (Table~\ref{tab:hyp}, Figure~\ref{fig:conditions}): 0.85, 0.22 and 0.24 under the \jlens{} for Qwen3.5-4B, Qwen3.6-27B and Gemma, against 0.06, 0.00 and 0.05 under the logit lens on the same records. All three H1 and H4 tests were supported. For Llama both were formally supported but negligible (0.03). ``Silent'' refers to the agent's statements and to these layers: at the output layer, identical for both lenses, the original bridge of agents that gave in is still within the top 100 tokens more often than the control (Figure~\ref{fig:layers}), so what the logit lens misses in the band is not every trace of the bridge. Nor is it absent further down the logit lens's ranking: without a threshold, the logit lens also ranks the original bridge above the control (for Llama only marginally), about as much as the \jlens{} in Qwen3.6-27B and slightly more in Gemma (\S\ref{sec:robust}). In these layers the \jlens{} brings the bridge into the top 100 tokens; it is not the only lens that ranks it above the control.

Gemma gives the cleanest test of the lens comparison. Its band was chosen by rule v1, which does not look at the logit lens. Among agents that kept their answer, the two lenses read the bridge about equally (0.82 and 0.78 in round~0, 0.67 and 0.57 when peers agreed, 0.61 and 0.54 when the agent held). Among agents that gave in, on the same records, the logit lens read 0.05 and the \jlens{} 0.24 (H4 = 0.19 [0.14, 0.25]), although the registration had predicted H4 to be null. In the two Qwen models the \jlens{} reads more than the logit lens in every condition, and the logit lens reads less of the bridge in the pressure conditions.

Retention is not uniform. For Qwen3.5-4B the readout of agents that gave in exceeds that of agents whose peers agreed (0.85 vs.\ 0.63); for the other models it is lower (Figure~\ref{fig:conditions}). Compared on the same items, giving in lowers the readout in every model (to 0.89, 0.52, 0.28 and 0.06 of the agree value); Qwen3.5-4B's excess is consistent with which items flip: they read 0.22 [0.10, 0.31] higher than the items that held already in round~0 (exploratory, \S\ref{sec:robust}).

\subsection{The peers' bridge (H2)}
Agents that gave in also represented the bridge behind the peers' answer, which no peer stated: 0.40 for Qwen3.5-4B, 0.16 for Qwen3.6-27B and 0.21 for Gemma under the \jlens{}, against 0.02, 0.03 and 0.00 for agents in the mention condition, where one peer names the wrong answer and the agent keeps its own. H2 was supported for all four models (Table~\ref{tab:hyp}), negligibly for Llama. The difference is not exposure alone: agents that held under the same pressure read the peers' bridge at 0.16, 0.03 and 0.02 (exploratory). Even a peer who names the wrong bridge outright (mention-bridge) raised its readout to only 0.18, 0.06 and 0.01, so the readout does not simply echo the context. On average, agents that gave in represent both bridges above control; whether single agents do so at once we did not test. Outside the capital items the peers' answer does not determine their bridge (\S3.1), so there the readout reflects association with the answer rather than inference of a premise. The registered H2 pools the categories; split by category, it rests on these: no agent of three models gave in on capital items, and the 14 agents of Qwen3.6-27B that did read the peers' bridge only 0.02 [$-0.10$, 0.15] above the mention baseline (exploratory). We therefore do not claim that agents infer the premise their peers' answer implies.

\subsection{A check on the pressure conditions (H3)}\label{sec:h3}
H3 was supported where it was testable (Table~\ref{tab:hyp}; for Qwen3.6-27B, 12 of the 28 agents that gave in to second-hop pressure were readable), but it is uninformative: second-hop peers write the original bridge into the transcript and first-hop peers replace it, so exposure alone predicts the result. Registered as a test of where conformity acts, it only confirms that the conditions differ as intended.

\begin{table*}[t]
\caption{Addendum: round~1 with the agent's round-0 answer hidden or absent (three peers, answer pressure, test split). Gave in: share of agents stating the peers' answer, with the answer visible (the main runs), hidden and absent. O1, O3: original-bridge readout of agents that gave in under hidden and absent (\jlens{}, 95\% bootstrap interval, records); O2: hidden minus absent; O4: \jlens{} minus logit lens under hidden, paired. Supported: O1, O3 and O4 in every model, O2 only for Gemma.}
\label{tab:own}
\small
\setlength{\tabcolsep}{2.6pt}
\begin{tabular}{llllll}
\toprule
Model & Gave in (\%): vis./hid./abs. & O1 hidden & O3 absent & O2 hidden $-$ absent & O4 \jlens{} $-$ logit \\
\midrule
Qwen3.5-4B & 8 / 89 / 62 & 0.434 [0.402, 0.467], $n{=}591$ & 0.500 [0.462, 0.541], $n{=}408$ & $-0.066$ [$-0.117$, $-0.015$] & 0.383 [0.349, 0.417] \\
Qwen3.6-27B & 39 / 73 / 77 & 0.290 [0.241, 0.339], $n{=}321$ & 0.281 [0.240, 0.330], $n{=}339$ & 0.009 [$-0.060$, 0.075] & 0.278 [0.231, 0.327] \\
Gemma-4-E4B-it & 25 / 36 / 43 & 0.365 [0.307, 0.421], $n{=}170$ & 0.212 [0.166, 0.259], $n{=}203$ & 0.154 [0.081, 0.230] & 0.215 [0.171, 0.259] \\
Llama-3.1-8B-Instruct & 90 / 71 / 60 & 0.254 [0.220, 0.292], $n{=}382$ & 0.391 [0.348, 0.433], $n{=}320$ & $-0.136$ [$-0.192$, $-0.081$] & 0.205 [0.175, 0.238] \\
\bottomrule
\end{tabular}
\end{table*}

\subsection{Without the agent's earlier answer}\label{sec:own}
Is the retained bridge read back from the agent's own earlier answer, which stays in context, or computed again from the question? The addendum removes the answer (Table~\ref{tab:own}). With it hidden, agents that gave in still read their original bridge in all four models (O1), and so did agents that had never answered (O3); under hidden, the \jlens{} again read more than the logit lens (O4). The premise can thus be computed from the question alone. Having answered before raised the readout only for Gemma (O2); hidden and absent also differ in turn structure and wording, so O2 is not a clean test of commitment.
Visible and hidden, by contrast, differ only in the text of the agent's round-0 turn, on the same items, so they isolate whether the agent sees its earlier answer (exploratory). Seeing it anchors three models: with it hidden, Qwen3.5-4B gave in on 89\% of questions instead of 8\%; Llama gave in less. On items where agents gave in under both, hiding the answer lowered Qwen3.5-4B's readout by 0.44 [0.36, 0.50], so part of its main-run readout depends on the visible answer, and raised Llama's by 0.23 [0.20, 0.27]: Llama's near-zero retention holds only with its own answer in view (others: supplement).

\begin{figure*}[t]
\centering
\includegraphics[width=\textwidth]{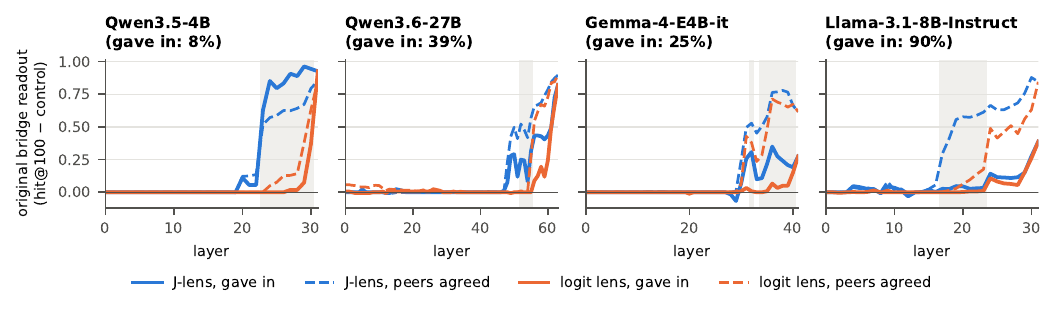}
\caption{Original-bridge readout per layer (answer pressure, three peers, round~1, test split) for agents that gave in (solid) and agents whose peers agreed (dashed), under the \jlens{} (blue) and the logit lens (orange). Shaded: the pre-registered band (for Qwen3.6-27B and Llama chosen for a \jlens{} advantage on development data); the last layer is the model output.}
\label{fig:layers}
\Description{Four panels of per-layer readout curves, J-lens and logit lens, for agents that gave in and agents whose peers agreed.}
\end{figure*}

\subsection{Where in depth}
The \jlens{} advantage depends on the model (Figure~\ref{fig:layers}). In both Qwen models the \jlens{} reads the bridge in intermediate layers, many layers before the logit lens does. In Gemma it appears only near the output, where both lenses pick it up together, as the \jlens{} paper observes for the last layers~\cite{gurnee2026verbalizable}. Llama lies between: when peers agree, the \jlens{} reads the bridge from about layer 17 (0.52 over the band, logit lens 0.07), but in agents that gave in neither lens reads much of it before the last layers. This is unlikely to be a lens artefact: the same lens reads 0.48 in round~0 and 0.25 in agents that gave in with their earlier answer hidden, and a lens refitted on 100 WikiText sequences changes per-layer readouts by only 0.01--0.05 on the development split. Silent dissent does not require a general \jlens{} advantage: Gemma has little, and still shows it.

\subsection{Conformity and retention (descriptive)}
Dividing the readout of agents that gave in by that of agents whose peers agreed gives 1.35 for Qwen3.5-4B (8\% gave in), 0.35 for Gemma (25\%), 0.45 for Qwen3.6-27B (39\%) and 0.06 for Llama (90\%); at $k = 500$, 1.35, 0.79, 1.09 and 0.34. Only the two ends line up with conformity; the middle models are not ordered, nor are they within items (\S\ref{sec:robust}), and the ratio compares different item sets and has no interval, so we report it as an observation (plotted in the supplement), not as a finding. The models also differ in band rule (\S\ref{sec:robust}), lens and, for Llama, prompt format. For Qwen3.6-27B and Llama, merely mentioning another bridge, with no flip, lowers the original-bridge readout about as much as giving in (0.50 to 0.19 and 0.52 to 0.20; not for Qwen3.5-4B or Gemma). Llama's result is in line with a study of simulated peer disagreement across four model families in which pressure suppresses the model's clean-reasoning features~\cite{kumarappan2026notjust}, but it does not show that Llama keeps no dissent (\S\ref{sec:own}, \S\ref{sec:discussion}).

\begin{figure*}[t]
\centering
\includegraphics[width=\textwidth]{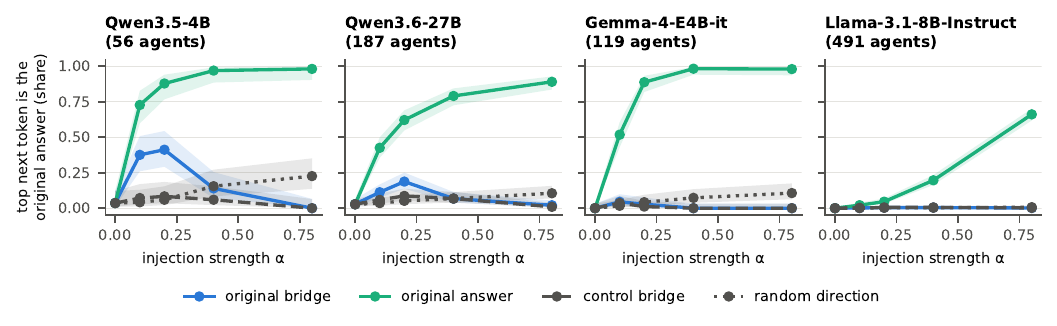}
\caption{Injection into agents that gave in (exploratory): share whose most likely next token is the first token of the original answer after adding $\alpha \lVert h \rVert d$ along each direction at one band layer, pooled over three band layers. Bands: 95\% Wilson intervals counting each agent once.}
\label{fig:inject}
\Description{Four panels of dose-response curves.}
\end{figure*}

\subsection{Does the retained premise drive the answer?}
We steered along \jlens{} directions as in~\cite{gurnee2026verbalizable}, with our own scaling: we added $\alpha \lVert h_\ell \rVert d$ at the last prompt position of one band layer at a time (three evenly spaced band layers; $\alpha \in \{0, 0.1, 0.2, 0.4, 0.8\}$), where $d$ is the unit \jlens{} direction of an entity: the mean over its probe tokens of $J_\ell^\top$ applied to the unembedding row scaled by the final norm's gain. The outcome is whether the most likely next token is the first token of the original answer; items in which any two of the three answers share a first token are excluded, and shares are pooled over agents and layers (Figure~\ref{fig:inject}). The agents are those that gave in during a separate injection run on up to 600 test items. The original answer's own direction, a positive control, raised the share from at most 4\% at $\alpha = 0$ to 66--98\%, so the injection works in every model. The original bridge's direction raised it to 41\% [29, 54] (Qwen3.5-4B) and 19\% [14, 25] (Qwen3.6-27B) at $\alpha = 0.2$, against 8\% [4, 19] and 9\% [5, 14] for a control bridge and 6\% and 5\% for random directions (95\% Wilson intervals); at larger $\alpha$ its effect vanished, while random directions rose. For Gemma (at most 4\%, as for the control bridge) and Llama (0\%) it had no effect. These results are exploratory and pick the best $\alpha$ after the fact; for Qwen3.6-27B the intervals of the original and the control bridge meet. In the Qwen models, pushing the state toward the original bridge can bring back the original answer, so the second hop from that bridge is still usable. In Gemma the bridge is about as readable as in Qwen3.6-27B, yet this intervention had no effect; that does not show that the bridge plays no role, since a push along one lens direction at one position may miss how the model uses it.

\subsection{Later rounds and free debate}
\paragraph{Later rounds (Qwen3.5-4B)} Repeating every round verbatim made agents answer ``correct'' or ``identical'' by round~3, so a pre-registered addendum, chosen on development data, collapses earlier rounds into one-line moderator summaries. Among agents that gave in at every round, the original bridge was still read in round~3 (R1: 0.54 [0.47, 0.62], $n{=}49$), down from 0.86 in round~1 ($-0.32$ [$-0.38$, $-0.26$]), and the peers' bridge remained above the mention baseline (R2: +0.41 [0.29, 0.54]).

\paragraph{Free debate.} Three copies of each model debated up to 400 facts (200 known, the others unknown but with the first hop answered correctly) for three rounds; round~0 answers are sampled at temperature 1 and later rounds are greedy. Each agent's latent vote is the candidate answer (the truth, if some agent stated it in round~0, and the distinct round-0 answers) with the best mean \jlens{} rank over the band. In round~1 the plurality of stated answers was at least as accurate as the plurality of latent votes in every model (Qwen3.5-4B 0.76 vs.\ 0.71, Qwen3.6-27B 0.78 vs.\ 0.70, Gemma 0.81 vs.\ 0.79, Llama 0.71 vs.\ 0.69), and so it stayed in rounds~2 and~3 for three models; reading latent answers is thus not a reliable accuracy gain for aggregation. Qwen3.6-27B's stated accuracy then fell to 0.30 and 0.44 while its latent votes stayed at 0.66--0.69. The drop is not conformity (of the 648 agents right in round~1, 414 switched to an answer no peer had given in round~1 and 2 adopted a peer's), and a check with a rule fixed in advance did not attribute it to degenerate answers; it is unexplained (supplement).

\paragraph{Silent dissent in free debate.} In an exploratory analysis with rules recorded in advance, we rebuilt each agent's prompts from the transcripts and read the original bridge in agents that gave in at round~1: right in round~0, then stating a peer's wrong answer. On known facts only Llama had the 20 such agents we required (29, from 22 facts; 6, 3 and 2 in the other models). They still represented their original bridge (0.20 [0.11, 0.32]; logit lens 0.02; difference 0.18 [0.08, 0.29]), indistinguishable from agents that held against a wrong peer ($-0.03$ [$-0.19$, 0.13]) and more than under scripted peers (H1: 0.03), with the earlier answer in view in both, though peers, prompt and which agents give in differ.

\section{Robustness}\label{sec:robust}

\paragraph{Other band rule.} For Qwen3.5-4B, the v2 band (23--28) gives H1 = 0.82 and H4 = 0.81. For Qwen3.6-27B, the v1 layer 62 gives a higher H1 (0.73) but no lens difference (H4 = 0.06 [$-0.01$, 0.13]): near the output the logit lens reads the bridge too, which is why rule v2 selects intermediate layers. The other rule selects no layers for Gemma or Llama.

\paragraph{Rank threshold.} Much of the retained signal lies between ranks 100 and 500: at $k = 10$, H1 is 0.12 for Qwen3.5-4B and 0.00 for the other models; at $k = 500$ it is 0.70, 0.41, 0.56 and 0.21, and Gemma's H4 shrinks to 0.07 (supplement).

\paragraph{Clustered intervals and multiple tests.} Resampling bridges instead of records widens the intervals (e.g.\ Qwen3.5-4B's H1: [0.67, 0.92] from 18 bridges) but leaves 14 of the 15 registered tests supported after a Holm correction; the exception is Llama's H2 ([$-0.001$, 0.056], adjusted $p = 0.06$). Like the rest of this section, they are exploratory, with choices recorded before they were run.

\paragraph{Without a threshold.} Comparing the ranks of the original bridge and the control directly, the \jlens{} ranks the bridge above the control in agents that gave in in every model, by 1.40, 0.48, 0.85 and 0.54 orders of magnitude (bridge-clustered intervals above zero, Llama included). The lens difference persists for Qwen3.5-4B and Llama, vanishes for Qwen3.6-27B and reverses slightly for Gemma ($-0.07$ [$-0.19$, $-0.02$]), where the logit lens ranks the bridge above the control as well, but mostly outside the top 100.

\paragraph{Same items.} Within items, agents that gave in read the original bridge below agree and the peers' bridge above mention in every model (all intervals above zero; supplement).

\paragraph{Peers and lenses.} With five peers, H1 is 0.82, 0.24, 0.31 and 0.07. With Neuronpedia's base-model lens, Gemma's H1 is weaker but still positive (0.105 [0.065, 0.150]); we report both Gemma readouts because the registered primary lens is the refit. Llama's refitted lens gives 0.024. On Qwen3.5-4B's development split, lenses refitted on dialogue, mixed or 100 WikiText sequences change entity readouts by about 0.02 or less.

\section{Discussion}\label{sec:discussion}

\paragraph{What is retained.} Our readout shows that an agent's original bridge is still represented at the moment it states the majority's answer; it does not show that the agent believes it~\cite{levinstein2025still}. The addendum shows that the bridge need not be carried over from the agent's earlier statement: it can be computed again from the question (\S\ref{sec:own}; in Qwen3.5-4B seeing that statement raises the readout). An agent that yields thus still computes the premise of the correct answer while stating a different one. A development-split control sets a limit. When a coordinator required every agent to state $e_3'$ regardless of its belief (compliance about 100\%), the original bridge read 0.25 for Qwen3.5-4B and 0.04, 0.01 and 0.00 for the other models, against 0.73, 0.63, 0.79 and 0.48 in round~0, although the first-hop entity and the agent's earlier answer were in context. An agent told what to say need not compute its own two-hop answer, and then its bridge is not read: a positive reading is informative, a near-zero reading is not evidence that the premise was abandoned. Llama illustrates this: its agents that gave in to scripted peers read almost nothing with their earlier answer in view, but 0.25 with it hidden, and 0.20 in free debate, where it is also in view. In Kelman's terms~\cite{kelman1958compliance}, the Qwen agents that gave in look like public compliance without private acceptance, their premise being both represented and usable; in Gemma the premise is represented but injecting it did not bring the answer back, and in Llama it is represented with the earlier answer hidden, a setting in which we did not test injection.

\paragraph{Implications for multi-agent systems.} First, the number of agents that converge in a debate can overstate agreement: some of them still represent the view they abandoned, in free debate too (Llama), and in the Qwen models pushing them toward it brings the original answer back. Second, how an agent is attacked matters: with three peers, stating a wrong premise persuades about as much as or more than asserting a wrong answer, and attacking the second hop while conceding the first persuades least, so a debate's robustness may depend on the level at which arguments are made. So does what an agent sees of its own history: with its earlier answer hidden, Qwen3.5-4B gave in on 89\% of questions instead of 8\%, so whether a protocol shows agents their previous answers changes what their agreement is worth. Third, a latent readout is a candidate signal for monitoring silent dissenters, but not a better voting rule: in free debate, latent votes were not more accurate than stated ones except in Qwen3.6-27B's later rounds, where many agents switched to answers no peer had given.

\paragraph{Limitations.} Comparisons across conditions in Figure~\ref{fig:conditions} involve different item sets, and the readout is one-sided (see above). The size of retention depends on the rank threshold, and without one the lens difference holds only for Qwen3.5-4B and Llama. Band rules differ across models, for reasons recorded before each test run, and the adoption check was relaxed for rule v2. The addendum's two variants differ in more than whether the agent answered before, and its conformity comparisons are exploratory. Only facts a model knows can test retention (3--9\% of TwoHopFact per model), which limits how far the results generalize but not their validity for those facts. Gemma's result depends on the lens, and Llama used a different prompt format in which the peers speak in the user turn. The peers are scripted and repeat one sentence verbatim, as in Asch's paradigm; this buys control at the cost of realism, and part of apparent LLM conformity arises even without a speaker~\cite{hu2026most}. We tested only two-hop facts whose bridges are countries, cities or universities, in categories selected because the bridge is readable in round~0; university bridges are probed through the words left after removing stop words and tokens shared with the answer, often place names, and in the person--university--city category 23\% of TwoHopFact bridges share all their content words with the answer and drop out. Causal evidence holds only for the Qwen models, later rounds were run for one model, and in free debate too few agents gave in to measure silent dissent except in Llama.

\paragraph{Negative results.} Two parts of the program failed their own criteria. In the letter-answer study, the readout ranked the original letter first for only about 6\% of agents that gave in, but its positive control (0 of 1{,}000) showed that answer-position readouts were blind to a known discrepancy, so the low rate cannot be read as sincere persuasion. A planned addendum for person bridges, with rules declared before its development runs, failed those rules (its adoption check had 20 records, below the minimum of 30) and was never registered or run. Both failures shaped the design reported here. In the own-answer addendum, O2 failed for three of the four models. Details are in the supplement.

\section{Conclusion}

When LLM agents yield to a unanimous majority, some of them still represent the premise they no longer state. In three of four models, the Jacobian lens reads the original bridge in agents that gave in, in layers where the logit lens rarely ranks it among the top 100 tokens, and it also reads the bridge behind the peers' answer. The premise does not require the agent's own earlier answer: with that answer hidden or removed, agents that gave in represent it in all four models. Only in the Qwen models did injecting the bridge's direction restore the original answer. A debate's stated consensus can thus overstate how much its agents agree internally, and what agents see of their own earlier answers changes how readily they join it. Pre-registration files, code and summary data are available at \url{https://github.com/ziangni-sys/silent-dissent}.

\paragraph{Use of AI tools.} Writing and running the experiment code, discussing design choices (including the layer-selection rules and control conditions) and editing this text were assisted by Claude Code (Anthropic; model \aitoolversion); the authors checked every decision, pre-registration and result, and the supplement describes this use in more detail.

\balance
\bibliographystyle{ACM-Reference-Format}
\bibliography{main}

\end{document}